# Modifiable Combining Functions*

Paul R. Cohen
*EKSL*
*Computer and Information Science*
*University of Massachusetts*
*Amherst, Massachusetts 01003*

Glenn Shafer
Prakash P. Shenoy
*School of Business*
*University of Kansas*
*Lawrence, Kansas 66045*

June 4, 1987

**Abstract**

Modifiable combining functions are a synthesis of two general approaches to combining evidence. Because they facilitate the acquisition, representation, explanation, and modification of expert knowledge about combinations of evidence, they are presented as a device for knowledge engineers, not as a normative theory of evidence combination. The basic idea of modifiable combining functions is to acquire degrees of belief for a subset of all possible combinations of evidence, then infer degrees of belief for other combinations in the set. If, in the course of knowledge engineering, a particular degree of belief is challenged, then it (and others) can be modified by an appropriate method.

## 1 Introduction

This paper presents a synthesis of two general approaches to combining evidence. When designing knowledge systems, knowledge engineers typically select one approach over the other, but each has strengths and weaknesses in terms of the ease with which knowledge can be acquired, represented, interpreted, modified, and explained. The synthesis we propose, called *modifiable combining functions* has many of the advantages of both approaches and overcomes some of their disadvantages. The basic idea of modifiable combining functions is to acquire degrees of belief for a subset of all possible combinations of evidence, then infer degrees of belief for other combinations in the set. If, in the course of knowledge engineering, a particular degree of belief is challenged, then it (and others) can be modified by an appropriate method.

A combination of evidence is a list of propositions, each with an associated degree of belief. For example, an expert system for diagnosing plant diseases has propositions like this:

$$((\text{soil texture} = \text{heavy}, .7)$$
$$(\text{soil oxygen} = \text{low}, .9))$$

That is, soil texture is believed to degree .7 to be heavy and soil oxygen is strongly believed (.9) to be low. Combinations of evidence are often found in the premises of inference rules. These rules can take two forms, called *specified* and *derived*:

---

*We thank Carole Beal, Sabine Bergler, Tom Gruber, and Adele Howe for their comments on drafts of this paper. This research is funded by a DARPA/RADC Contract F30602-85-C-0014.

specified form:

```
IF      ((soil texture = heavy,  .7)
         (soil oxygen = low,  .9))
THEN    (water damage = yes,  .8)
```

derived form:

```
IF      ((soil texture = heavy,  x)
         (soil oxygen = low,   y))
THEN    (water damage = yes,  f (x, y, k))
```

These forms suggest two general approaches to combining evidence. The specified form requires that for each combination of degrees of belief in the premise, a degree of belief is specified for the conclusion. The derived form requires a function, $f$, that derives a degree of belief in the conclusion for *any* degrees of belief in the premise. The constant k in the derived form represents the degree of belief that would be assigned to the conclusion if the degree of belief in the premise was 1.0, that is, the degree of belief in the inference rule itself. This quantity is implicit in the specified form.

These forms combine evidence *within* inference rules, but they have counterparts for the cases in which two or more rules draw the same, corroborating conclusion. By analogy with the specified form, degrees of belief can be acquired for each combination of corroborating rules; or a general function, analogous to $f$ in the derived form, can be acquired to calculate degrees of belief for all corroborations.

Both approaches have been used in AI systems. Considering medical expert systems alone (reflecting our own interest in this area), we find knowledge in the specified form in PIP (Pauker, Gorry, Kassirer, & Schwartz, 1976), IRIS (Trigoboff, 1978), MDX (Chandrasekeran, Mittal, & Smith, 1982), and MUM (Cohen et al., 1986); while MYCIN (Shortliffe, 1976; Shortliffe, Buchanan, 1975), INTERNIST/CADUCEUS (Pople, 1977) and ABLE (Patil, Szolovits, & Schwartz, 1976) use knowledge in the derived form.

In outline, we describe representations for combining functions that are closely-related to the specified and derived forms. We will discuss the tradeoffs between these approaches that motivate the idea of modifiable combining functions. We will illustrate how modifiable combining functions are generated and modified in the context of an example. Parts of the theory of modifiable combining functions have been implemented in a medical expert system (Cohen et al., 1986), but much of this paper should be taken as research in progress.

## 2  Forms of combining functions

### 2.1  Tabular combining functions

Tabular combining functions are often represented as tables that specify degrees of belief in conclusions for each combination of degrees of belief of evidence. Figure 1 shows a tabular function that combines two pieces of evidence, $E_1$ and $E_2$, for conclusion C. In this case, degrees of belief in evidence range from -1 to +1, denoting complete disbelief and belief, respectively. A degree of belief of zero denotes ignorance; for example (soil oxygen = low, 0) means that the value of soil oxygen is unknown, either because it is an unavailable datum or because the data from which it is inferred are ambiguous. Many of the cells are blank, meaning that the expert does not consider these combinations of evidence relevant – does not expect them to arise during problem solving. From the knowledge engineer's perspective, blank cells and zero cells represent different events. A blank means that that particular combination of evidence was never considered, but a zero means it was considered and found to be uninformative. From the perspective of an AI program's interpreter, blank and zero *may* both mean that the combination of evidence is uninformative; or a blank may be used to alert the user to incompleteness in the combining function.

In tabular combining functions, degrees of belief in evidence index degrees of belief in conclusions. The combining function in Figure 1 specifies that when the degrees of belief in $E_1$ and $E_2$ are .5 and .75, respectively, the degree of belief in C is .25. Since conclusions are often used as evidence for subsequent inferences, the cells in tabular combining functions contain values that can themselves be used to index degrees of belief in other tabular functions. Tabular functions increase exponentially in size: A function for N pieces of evidence requires an N-dimensional table, similar to the *signature tables* invented by Samuel (1959).

Some important knowledge about patterns or regularities in combinations of evidence is implicit in tabular combining functions. For example, the entire upper-right quadrant of Figure 1 is blank, suggesting that no combination of positive degrees of belief in $E_1$ and negative degrees of belief in $E_2$ is meaningful. Similarly, in the lower-left quadrant we see a *threshold* on the degree of belief in $E_1$: the values in the table are determined by $E_2$ for all values of $E_1$ less than or equal to -.75. These regularities are easily captured by a *rule-based variant* of tabular combining functions. The two examples we just mentioned can be represented this way:

**Upper-right quadrant:**

$$\begin{aligned} \text{IF bel}(E_1) &\geq 0 \text{ and} \\ \text{bel}(E_2) &\leq 0 \\ \text{Then bel}(C) &:= 0 \end{aligned}$$

**Lower-left quadrant:**

$$\begin{aligned} \text{IF bel}(E_1) &\leq -.75 \text{ and} \\ \text{bel}(E_2) &= .5 \text{ or} \\ \text{bel}(E_2) &= .25 \\ \text{Then bel}(c) &:= -.75 \end{aligned}$$

$$\begin{aligned} \text{IF bel}(E_1) &\leq -.75 \text{ and} \\ \text{bel}(E_2) &= .75 \\ \text{Then bel}(c) &:= -.5 \end{aligned}$$

Irrespective of whether the knowledge engineer acquires tables like Figure 1, or rules as above, he or she must take care to maintain important distinctions in the domain. For example, the rule for the upper-right quadrant could be extended to account for the blank cells in the lower-right quadrant, too, by changing its first clause to "IF bel($E_1$) $\geq$ -.5." While this rule describes the table, it obscures what may be an important distinction between positive and negative values for $E_1$.

Tabular combining functions and their rule-based variant are ways to represent combinations of evidence given in the specified form, described above. A representation that relies on both specified and derived combinations is discussed next.

## 2.2 Interpolated combining functions

Three of the four corner cells of Figure 1 represent degrees of belief in the conclusion given categorical (certain) data about $E_1$ and $E_2$ (the upper-right cell is blank because nothing is known about it.) They can be arranged in a *categorical table* as shown below. To distinguish categorical tables from the larger ones like Figure 1, we call the latter *full* tables.

$$\text{bel}(E2): \begin{array}{c|c|c|} & \multicolumn{2}{c}{\text{bel}(E1):} \\ & 1 & -1 \\ \hline 1 & 1 & X \\ \hline -1 & 0 & -1 \\ \hline \end{array}$$

The upper-left cell contains the degree of belief in C given that $E_1$ and $E_2$ are both true; conversely, the lower-right cell is the degree of belief in C when both are false; the 0 in the lower-left cell represents ignorance in C given that $E_1$ is true and $E_2$ is false. To reiterate, these are the corner cells of the full table in Figure 1. All other, noncorner cells in Figure 1 represent interpolations between the the values in this categorical table, interpolations due to uncertainty in $E_1$ and $E_2$. For example, the cells around the center of Figure 1 tend toward the value 0, since the center cell represents the case in which the degrees of belief in $E_1$ and $E_2$ are both zero, that is, completely uninformative. Similarly, in the lower half of the table, we see degrees of belief in C ranging from 0 when $\text{bel}(E_2) = +1$, to -1 for lower degrees of belief in $E_2$.

The full table in Figure 1 was built by hand, but full tables can also be derived by interpolating functions. Figure 2 shows the derivation of a full table by a Bayesian interpolating function. The categorical corner cells are 1.0, .95, .25, and 0.0, respectively. All other cells contain intermediate values that reflect uncertainty about the evidence. For example, when the degrees of belief in *episode* and *risk factors* are both .75, the degree of belief in the conclusion is .79, a value intermediate between the four corner points but nearer to 1.0 — its nearest neighbor — in magnitude. This table and its derivation will be explained in Section 5.

To summarize, full tables can be built by hand, by specifying the value in each cell, or specifying rules that assert the values of subsets of the cells. Alternatively, they can be derived automatically by interpolating from categorical tables. Once the decision has been made to use interpolating functions, full tables are usually not generated and stored. Instead, the values of combinations of evidence are computed as needed. However, the following Section suggests that there are advantages to keeping both forms of combining functions.

## 3 Comparison

Our comparison will focus on the tabular and interpolating forms of combining functions. The strengths of one often correspond to weaknesses in the other. First, tabular combining functions do not infer anything that is not stated by the expert. Most of the cells in a table are blank, meaning that the expert does not consider them to represent meaningful combinations of evidence. In theory, every nonblank cell represents a meaningful combination and every blank cell represents a meaningless one. But in practice, the sheer size of tabular functions means that some meaningful combinations of evidence are simply overlooked during knowledge acquisition. In this sense, tabular functions are *brittle*: they cannot account for all meaningful situations that will arise during problem solving.

Interpolation is clearly a solution to the brittleness problem, since the value of any blank cell can be inferred from the corners of a categorical table, or perhaps from its "nearest neighbors." The disadvantages of interpolating functions are that, unlike tabular functions, they produce values for *all* combinations of evidence in their domain, meaningful or not. Moreover, no value derived by an interpolating function is guaranteed to reflect an expert's judgment. A subtler problem is that interpolation produces a *continuous* gradient of values between the corners of the full table. But expert's degrees of belief in conclusions are unlikely to change continuously with the degrees of belief in the evidence. Thresholds are common, as illustrated by the rule-based variant of tabular functions.

Tabular functions are *locally modifiable,* meaning that a knowledge engineer can change the values of individual cells in the table with the assurance that the performance of the system will remain unchanged except in the cases of these particular combinations of evidence. This allows a combining function to be "tuned" in the normal course of knowledge base refinement: when the system presents a conclusion that the expert thinks is wrong, and the source of the error is localized to a particular cell, then that cell can be changed. In contrast, changing an interpolating function necessarily effects the values assigned to *all* combinations of evidence in its domain. Modifying an interpolating function is essentially redesigning one's inference system (Gruber and Cohen, 1987).

## 4  Modifiable Combining Functions

Once the knowledge engineer considers using interpolating functions, why bother to acquire full tables by hand? Why not simply acquire categorical tables, as above, and design interpolating functions to, in effect, "fill in" the intermediate values? Clearly, the two approaches are equivalent if the interpolating functions generate the same values as the expert for any combination of evidence. But there is no way to test this, other than to acquire an entire table and then compare it with the results of an interpolation function. Consequently, the knowledge engineer can take one of two positions with respect to potential differences between interpolated values and the expert's judgment:

- The knowledge engineer can design a function that has desirable properties and assume that, if the expert's judgment is different, it is because the expert's reasoning is inconsistent or otherwise flawed.

- The knowledge engineer can design a function that is assumed to reflect expert judgment, but modify it to conform to the expert when deviations become apparent.

The first position is associated with normative models, the second with performance models. In both cases, the knowledge engineer must carefully design interpolation functions given what he knows and can assume about the evidence in a domain. In the latter case, in addition, he must have some mechanism for modifying combining functions.

Modifiable combining functions are a synthesis of tabular and interpolating functions. They are tabular functions that have most of their values derived by interpolation, but that can be modified to conform to an expert's judgment. Knowledge engineers must first acquire a categorical table and any other cells in the full table that the expert can provide. Interpolating functions ideally should fill in cells that the expert and knowledge engineer neglected to specify, with values that are likely to match the expert's judgment, but not fill in cells they intended to leave blank. If these goals are not achieved, the tabular function can be modified by one of the three mechanisms discussed below.

## 5  An Example

This section illustrates modifiable combining functions for two pieces of evidence from a medical diagnosis problem. Most diagnosis begins with the physician taking a *history*: asking about the patient's chief complaint, age, past medical history, and so on. Our example concerns the diagnosis of *angina* and two pieces of evidence from the history: the patient's report of an *episode* of chest pain, and whether the patient has *risk factors* for angina. Clearly, other evidence plays a role in diagnosis, but we will focus on a single rule that infers that the patient's history is consistent with angina if he or she has a characteristic episode and risk factors:

$$episode\ \&\ risk\ factors\ \rightarrow\ angina\ history$$

Both pieces of evidence can be uncertain because each depends on several observations. For the purpose of this example, assume that degrees of belief in *episode* and *risk factors* are subjective probabilities ranging from 0.0 to 1.0. The interpretation of P(*episode*) = 0 is "the episode is not characteristic of angina." An intermediate degree of belief, say P(*episode*) = .5, means "some aspects of the episode are consistent with angina, but other aspects are missing." The following examples illustrate assessments of degrees of belief for particular observations:

- crushing chest pain, induced by exercise, lasting a few minutes, radiating to one or both arms, accompanied by sweating and shortness of breath: P(*episode*) = 1.0

- sharp, fleeting chest pain, induced by sudden movement, not radiating: P(*episode*) $\doteq$ 0.0

- diffuse chest pain, came on after eating, radiating, lasting about 30 seconds: P(*episode*) = 0.5

- 60 year-old male, overweight, smoker, with high blood pressure, and two brothers with coronary artery disease: P(*risk factors*) = 1.0

- 30 year-old female, nonsmoker, not overweight, normal blood pressure, no history of heart disease in the family: P(*risk factors*) = 0.0

- 45 year-old male, smoker, not overweight, marginally-high blood pressure, uncle had coronary at age 60: P(*risk factors*) = .5

Given that *episode* and *risk factors* can be uncertain, how should a knowledge engineer acquire knowledge about the combinations of this evidence that support (or detract from) the conclusion? Degrees of belief for all possible combinations could be acquired in the *specified* form, and arranged in a tabular combining function. Alternatively, the knowledge engineer might design a combining function, $f$, and derive the degrees of belief of combinations by interpolation.

Modifiable combining functions present an intermediate alternative: the knowledge engineer acquires some degrees of belief for a subset of the possible combinations, then designs a function to interpolate the values of the rest and arranges the results in a table, then modifies the table if necessary to accord with the expert's judgment. An obvious place to begin this process is with the categorical table, from which a full table can be interpolated. Imagine the following rules, qualified by degrees of belief, are acquired from the expert:

*episode* & *risk factors* $\rightarrow$ *angina history* , 1.0
*episode* & $\sim$ *risk factors* $\rightarrow$ *angina history* , .95
$\sim$ *episode* & *risk factors* $\rightarrow$ *angina history* , .25
$\sim$ *episode* & $\sim$ *risk factors* $\rightarrow$ *angina history* , 0.0

These can be arranged in the following categorical table:

|  | P(episode) | |
|---|---|---|
|  | 1 | 0 |
| P(risk factors): 1 | 1.0 | .25 |
| P(risk factors): 0 | .95 | 0.0 |

The knowledge engineer needs to design a function from which P(*angina history*) can be derived for values of P(*episode*) and P(*risk factors*) other than 1 and 0. Such functions reflect the knowledge engineer's assumptions about the domain. We will illustrate a Bayesian function designed under the assumption that P(*episode*) and P(*risk factors*) are independent.

The Bayesian interpolation function is derived from the rule of total probability, which says

15

$$P(A) = \sum_{i=1 \to n} P(A|B_i)P(B_i)$$

where $B_1, \ldots, B_n$ is an exhaustive list of mutually exclusive possibilities. For our example, A is the conclusion *angina history* and $B_1, \ldots, B_n$ is

*episode & risk factors*
*episode & ~ risk factors*
*~ episode & risk factors*
*~ episode & ~ risk factors*

Then, P(*angina history*) can be derived for any degrees of belief in *episode* and *risk factors* as follows:

P(a) =
P(a | e & r) P(e & r)
+ P(a | e & ~ r) P(e & ~ r)
+ P(a | ~ e & r) P(~ e & r)
+ P(a | ~ e & ~ r) P(~ e & ~ r)

where *episode, risk factors* and *angina history* are abbreviated e, r, and a, respectively.

The values of the conditional terms in this expression have already been acquired from the expert and are recorded in the categorical table (e.g., $P(a|e \& r) = 1.0, P(a|e \& \sim r) = .95\ldots$). The knowledge engineer now must decide whether to acquire the other terms in the expression P(e & r), P(e & ~r) ... This effort can be avoided by assuming that e and r are independent, in which case P(e & r) = P(e)P(r), and

P(a) =
P(a | e & r) P(e)P(r)
+ P(a | e & ~ r) P(e)P(~ r)
+ P(a | ~ e & r) P(~ e)P(r)
+ P(a | ~ e & ~ r) P(~ e)P(~ r)

or,

P(a) =                                                                                    (1)
P(a | e & r) P(e)P(r)
+ P(a | e & ~ r) P(e)[1 - P(r)]
+ P(a | ~ e & r) [1 - P(e)]P(r)
+ P(a | ~ e & ~ r) [1- P(e)][1 - P(r)] .

Figure 2 illustrates a full table containing the values of P(a) derived by this function from these categorical values

P(a | e & r) = 1.0
P(a | e & ~ r) = .95
P(a | ~ e & r) = .25
P(a | ~ e & ~ r) = 0

and letting P(e) and P(r) range through the values 0, .125, .25, .375, .5, .625, .75, .825, and 1.0.

The Bayesian function (1) is an example of what is sometimes called Jeffrey's rule (Shafer, 1981; Shafer and Tversky, 1985, p.333). In such a design the conditional probabilities $\dot{P}(a|e\&r)$ ... reflect the expert's heuristic judgments based on previous cases of angina. In contrast, the unconditional probabilities $P(e\&r) = P(e)P(r)$ ... reflect knowledge about the individual patient who is currently being diagnosed. This is because, to calculate $P(e\&r)$, we assume that the probability of an angina episode is independent of whether one is at risk. This is true for an individual patient: for *this* patient the probability of an angina episode *is* independent of the probability that he is at risk. *This* patient either has risk factors in addition to his angina episode or he doesn't. Thus, the decision to design a Bayesian function for which $P(e\&r) = P(e)P(r)$ implies that the expert's knowledge of patients in general is dominated by his knowledge of the probabilities $P(e)$ and $P(r)$ for the individual patient.

Consider how this assumption might lead to a conflict with the expert's judgment. In general,

$$P(e \ \& \ r) = P(e|r)P(r)$$

or, if $P(e)$ and $P(r)$ are independent, then $P(e|r) = P(e)$ and

$$P(e \ \& \ r) = P(e)P(r)$$

But this seems wrong because, in general, the probability of an episode given risk factors is higher than the probability of the episode, or

$$P(e|r)P(r) > P(e)P(r)$$

Consequently, in some cases, $P(e \ \& \ r)$ will be too low, and so the value of $P(a)$ denoted by (2) will be too low, as well. For example, according to Figure 2, if $P(episode) = .5$ and $P(risk\ factors) = .75$, then $P(angina\ history) = .59$. But in the course of testing a system, the expert may challenge this result. He may say that if there is moderate evidence of an episode and strong evidence of risk factors then the probability of angina history should be much higher, say, 0.75.

What should the knowledge engineer do in this case? If he is relying exclusively on interpolating functions then he has 3 options:

1. insist that the expert's judgment is flawed

2. change the categorical table

3. change the interpolating function

The first is practical only if the knowledge engineer is confident that the assumptions that underlie his interpolating function are reasonable. The other two have global effects on all the numbers in the table, not just the few the expert criticized. Thus, in fixing the immediate problem the knowledge engineer could introduce new ones. Knowledge engineering often extends over a period of months, and the knowledge engineer relies on a kind of monotonicity — the idea that adding new knowledge to a system will not make it perform differently on the majority of previous cases. Changing the categorical table has ramifications only for the inference rule with which it is associated, but changing an interpolation function will change the degrees of belief of all the conclusions derived by that function — potentially every conclusion previously derived by a knowledge system.

If the knowledge engineer *does* decide to change the function, how should he go about it? We could change (1) by eliminating the independence assumption and acquiring the required conditional probabilities from the expert. Or, we might design a completely new Bayesian function that exploits the causal associations between the evidence and the conclusion (Pearl, 1986). Or, we could conclude that a belief-function design better characterizes the relationship between the evidence and the conclusion (Shenoy and Shafer, 1986)[1]. Many interpolation schemes are possible, but most of them are mathematically complicated, or computationally expensive, or require many more numbers than the expert can

---

[1] The authors are currently working on a formulation of modifiable combining functions based on belief functions. The formulation is preliminary, and space limitations preclude introducing it here.

17

accurately provide. The Bayesian function above (1) is very simple and requires few numbers. Its major deficit is that, in a few cases, it produces numbers with which the expert disagrees.

If the knowledge engineer does not rely exclusively on interpolating functions to calculate degrees of belief, then he has another option besides the three listed above: He can simply change the values that the expert says are wrong and store the new values in a tabular form that overides the derived values. The idea of modifiable combining functions is, in essence, to use simple interpolating functions to derive full tables from categorical tables, then, when the expert criticizes a derived degree of belief, to simply change it. This is shown in Figures 3 and 4. In Figure 3, the expert identifies a block of cells with values that are too low, for the reasons we discussed earlier. Figure 4 shows one possible modification.

In sum, modifiable combining functions offer three methods for representing expert judgments about combinations of evidence. First, individual cells, or blocks of cells in a derived tabular function can be changed. Second, the value in the categorical table can be changed. Third, and as a last resort, the interpolating function can redesigned.

## 6 Conclusion

Modifiable combining functions share many of the advantages of tabular and interpolating functions while avoiding some of their disadvantages. The information burden of tabular functions is reduced because the full table is derived by interpolating from the values the expert *can* provide. (One natural basis for the interpolation is the categorical table, but others are possible.) The brittleness of tabular combining functions, especially multidimensional ones, is overcome. Simple interpolating functions can be used, requiring relatively few numbers from the expert. Then, any values in the derived full table can be overridden by the expert's judgment. Discontinuities can easily be expressed in the rule-based variant of tabular combining functions. When an interpolating function fills in cells that the expert thinks should be blank (meaningless), the function can be modified accordingly. All modifications to cells are local in the sense that they affect the system's performance for combinations of evidence represented by those cells only. But if global modifications are appropriate, if *all* the values in a modifiable combining function seem wrong to the expert, then the knowledge engineer can first consider modifying the categorical table (or any other set of points used for interpolation) and then consider modifying the interpolation function.

Currently, we are acquiring tabular combining functions for a medical expert system (Cohen, Day, Delisio, Greenberg, Kjeldsen, Suthers, and Berman, 1986) and a plant pathology system. They are represented as rules, as discussed above. We have built interfaces for acquiring and modifying these rules, and we have almost completed a graphic interface for representing them in tabular form. Currently, we do not fill in the values of empty cells by interpolation, so the full promise of modifiable combining functions has yet to be demonstrated.

18

Categorical Table

| 1.0 | .25 |
|---|---|
| .95 | 0 |

bel($E_1$)

| bel($E_2$) | 1.0 | .75 | .50 | .25 | 0 | -.25 | -.50 | -.75 | -1.0 |
|---|---|---|---|---|---|---|---|---|---|
| 1.0 | 1.0 | 1.0 | .75 | .50 | 0 | | | | |
| .75 | 1.0 | 1.0 | .50 | .25 | 0 | | | | |
| .50 | .50 | .25 | 0 | 0 | 0 | | | | |
| .25 | .25 | 0 | 0 | 0 | 0 | | | | |
| 0 | 0 | 0 | 0 | 0 | 0 | | | | |
| -.25 | | | | | | | | | |
| -.50 | | | | | | | | | |
| -.75 | 0 | -.50 | -.75 | -.75 | -1 | -1 | -1 | -1 | -1 |
| -1.0 | 0 | -.50 | -.75 | -.75 | -1 | -1 | -1 | -1 | -1 |

Figure 1

P(episode)

| P(risk factor) | 1.0 | .875 | .75 | .625 | .50 | .375 | .25 | .125 | 0 |
|---|---|---|---|---|---|---|---|---|---|
| 1.0 | 1.0 | .91 | .81 | .72 | .63 | .53 | .44 | .34 | .25 |
| .875 | .99 | .90 | .80 | .70 | .61 | .51 | .41 | .32 | .22 |
| .75 | .99 | .89 | .79 | .69 | .59 | .49 | .39 | .29 | .19 |
| .625 | .98 | .88 | .78 | .67 | .57 | .47 | .36 | .26 | .16 |
| .50 | .98 | .87 | .76 | .66 | .55 | .44 | .34 | .23 | .13 |
| .375 | .97 | .86 | .75 | .64 | .53 | .42 | .31 | .20 | .09 |
| .25 | .96 | .85 | .74 | .63 | .51 | .40 | .29 | .18 | .06 |
| .125 | .96 | .84 | .73 | .61 | .49 | .38 | .26 | .15 | .03 |
| 0 | .95 | .83 | .71 | .59 | .48 | .36 | .24 | .12 | 0 |

Figure 2

Categorical Table

| 1.0 | .25 |
|---|---|
| .95 | 0 |

P(episode)

| P(risk factor) | 1.0 | .875 | .75 | .625 | .50 | .375 | .25 | .125 | 0 |
|---|---|---|---|---|---|---|---|---|---|
| 1.0 | 1.0 | .91 | .81 | .72 | .63 | .53 | .44 | .34 | .25 |
| .875 | .99 | .90 | .80 | .70 | .61 | .51 | .41 | .32 | .22 |
| .75 | .99 | .89 | .79 | .69 | .59 | .49 | .39 | .29 | .19 |
| .625 | .98 | .88 | .78 | .67 | .57 | .47 | .36 | .26 | .16 |
| .50 | .98 | .87 | .76 | .66 | .55 | .44 | .34 | .23 | .13 |
| .375 | .97 | .86 | .75 | .64 | .53 | .42 | .31 | .20 | .09 |
| .25 | .96 | .85 | .74 | .63 | .51 | .40 | .29 | .18 | .06 |
| .125 | .96 | .84 | .73 | .61 | .49 | .38 | .26 | .15 | .03 |
| 0 | .95 | .83 | .71 | .59 | .48 | .36 | .24 | .12 | 0 |

Figure 3

Categorical Table

| 1.0 | .25 |
|---|---|
| .95 | 0 |

P(episode)

| P(risk factor) | 1.0 | .875 | .75 | .625 | .50 | .375 | .25 | .125 | 0 |
|---|---|---|---|---|---|---|---|---|---|
| 1.0 | 1.0 | .91 | .81 | .75 | .75 | .50 | .50 | .34 | .25 |
| .875 | 99 | .90 | .80 | .75 | .75 | .50 | .50 | .32 | .22 |
| .75 | .99 | .89 | .79 | .75 | .75 | .50 | .50 | .29 | .19 |
| .625 | .98 | .88 | .78 | .75 | .75 | .50 | .50 | .26 | .16 |
| .50 | .98 | .87 | .76 | .66 | .55 | .44 | .34 | .23 | .13 |
| .375 | .97 | .86 | .75 | .64 | .53 | .42 | .31 | .20 | .09 |
| .25 | .96 | .85 | .74 | .63 | .51 | .40 | .29 | .18 | .06 |
| .125 | .96 | .84 | .73 | .61 | .49 | .38 | .26 | .15 | .03 |
| 0 | .95 | .83 | .71 | .59 | .48 | .36 | .24 | .12 | 0 |

Figure 4